\documentclass[11pt,a4paper]{article}

\usepackage{amsmath,amssymb,amsfonts,amsthm}
\usepackage{algorithm, algpseudocode}
\usepackage{graphicx}
\usepackage{textcomp}
\usepackage{xcolor}
\usepackage{tikz}
\usepackage{pgfplots}
\usetikzlibrary{external, positioning,shapes, arrows}
\usepackage{mathtools}
\usepackage{varwidth}
\usepackage[utf8]{inputenc}
\usepackage[T1]{fontenc}
\usepackage{enumitem}
\usepackage{siunitx}
\usepackage{balance}
\usepackage{multirow}
\usepackage{booktabs}

\usepackage[backend=biber, style=ieee, sorting=none, dashed=false]{biblatex}

\usepackage{fancyhdr}
\fancypagestyle{FirstPage}{
	\cfoot{\footnotesize\textbf{Original publication DOI:} 10.1109/DCOSS52077.2021.00061\\ \textbf{\copyright2022 IEEE.} Personal use of this material is permitted. Permission from IEEE must be obtained for all other uses, in any current or future media, including reprinting/republishing this material for advertising or promotional purposes, creating new collective works, for resale or redistribution to servers or lists, or reuse of any copyrighted component of this work in other works.} 
}
\begin{document}
\newbox\keywbox
\setbox\keywbox=\hbox{\bfseries Keywords:}%

\newcommand\keywords{%
\noindent\rule{\wd\keywbox}{0.25pt}\\\textbf{Keywords:}\ }

\title{Trajectory-based Traveling Salesman Problem for Multirotor UAVs}
\author{$\text{Fabian Meyer}^{1\text{st}}$, fabian.meyer@fzi.de \and $\text{Katharina Glock}^{2\text{nd}}$, kglock@fzi.de}
\date{}
\maketitle

\begin{abstract}
	This paper presents a new method for integrated time-optimal routing and trajectory optimization of multirotor unmanned aerial vehicles (UAVs). Our approach extends the well-known Traveling Salesman Problem by accounting for the limited maneuverability of the UAVs due to their kinematic properties. To this end, we allow each waypoint to be traversed with a discretized velocity as well as a discretized flight direction and compute time-optimal trajectories to determine the travel time costs for each edge. We refer to this novel optimization problem as the Trajectory-based Traveling Salesman Problem (TBTSP). The results show that compared to a state-of-the-art approach for Traveling Salesman Problems with kinematic restrictions of UAVs, we can decrease mission duration by up to 15\%.
\keywords Traveling Salesman Problem, multirotor UAV, route planning, trajectory planning, time-optimality
\end{abstract}

\section{Introduction}
\thispagestyle{FirstPage}
In recent years, unmanned aerial vehicle (UAV) technology has been steadily gaining momentum. With technological advancements, UAVs are proving to be extremely useful in a variety of application scenarios. These include monitoring and inspection of large infrastructures and energy facilities such as offshore wind farms, power lines, roads, oil and gas pipelines \cite{Beseda.2018, Mader.2016, Keller.2017, Otto.2018}, monitoring of cultivated land and forests \cite{Otto.2018, DHL.2014}, in the mining industry \cite{Shahmoradi.2020}, layout planning and digital reconstruction in construction \cite{Zheng.2018} and for damage assessment after disaster events \cite{Restas.2015, Glock.2020}.

To perform aerial flights in the above-mentioned use cases, two aspects are of crucial importance. On the one hand, the individual waypoints of a mission have to be put in a suitable order to minimize unnecessary time and energy consumption. This aspect is covered by solving route planning problems such as the NP-hard Traveling Salesman Problem (TSP), which is discussed in detail in \cite{Laporte.1992}. On the other hand, there is UAV motion planning, which connects two waypoints of the respective missions in a way that the movement between them is compatible with the physical properties of the UAV. Motion planning itself divides into two different subfields namely trajectory optimization and path planning. While path planning exclusively calculates the shortest feasible spatial path between two waypoints, trajectory planning aims to find the fastest. Until a few years ago, route planning and motion planning developed almost separately from each other, although they are mutually dependent.

Approaches to combine the research field of motion planning with route planning are based predominantly on path planning methods. An example are so-called Dubins paths, which are based on the principle of minimum turning radii (see \cite{Otto.2018}). A disadvantage of these methods, however, is that they are often based on the assumption of a constant velocity and thus cannot be used universally. Especially for the mission planning of the widely used multirotor UAVs, the degree of freedom offered by longitudinal acceleration, which makes them so flexible to use, is thus omitted.

In this paper, we present a new concept for mission planning of multirotor UAVs, which can fully exploit their physical properties. The problem we solve in this work combines the well known Traveling Salesman Problem with time-optimal trajectory generation and is therefore called the Trajectory-based Traveling Salesman Problem (TBTSP). Possible applications include surveying and monitoring of static objects such as buildings and infrastructures, where flight patterns can be computed offline in advance. The contributions of this paper are as follows.

\begin{itemize}
	\item Presentation of the TBTSP model for UAV route planning that explicitly allows discretized velocities and heading angles in each waypoint to be visited.
	\item A concept for combining the TBTSP model with a time-optimal trajectory optimization method to determine flight trajectories in an environment without obstacles, taking into account the UAV's maximum velocity and acceleration.
	\item A computational study and benchmarking of our proposed approach against exact solutions of the discretized Dubins TSP (DDTSP) as state-of-the-art approach.	
\end{itemize}

This paper is organized as follows. Section \ref{Sec:RelatedWork} gives an overview of the state-of-the-art. This mainly covers the approaches for UAV route planning problems that already consider motion planning, followed by UAV trajectory optimization methods. In Section \ref{Sec:Routing}, the TBTSP model for UAV route planning considering variable velocities and heading angles is presented. Next, Section \ref{Sec:TrajectoryGeneration} presents the trajectory generation method used in this work. A computational study can be found in Section \ref{Sec:results} and finally, a conclusion and outlook are given in Section \ref{Sec:Conclusion}.

\section{Related Work}\label{Sec:RelatedWork}
Route planning problems for UAVs are enjoying growing interest in the scientific literature. While algorithms for solving UAV route planning problems predominantly used simple Euclidean distances for time-of-flight estimation a few years ago, the trend has moved towards more precise and practical feasible metrics for time-of-flight estimation \cite{Henchey.2020}.
 
  To do this, established methods include certain aspects of UAV movements into route planning. One of these aspects is the minimum turning radius. This is the maximum curvature of the trajectory of a UAV moving at constant velocity and maximum lateral acceleration. It is a restriction that comes into effect especially for fixed-wing UAVs but is also considered for multirotor UAVs.

The resulting flight paths are known as Dubins paths. Dubins showed in his work \cite{Dubins.1957} that the shortest path between two points of the plane for which start and end tangents are fixed and which does not exceed a maximum curvature always consists of three segments representing either a left or right curve or a straight line.  This principle is used in many UAV route planning problems. Examples here include \cite{Cohen.2017, Penicka.2017}, and \cite{Geng.2013}. The advantage of using Dubins paths and thus assuming constant velocity is that acceleration and deceleration are avoided, making the trajectory more energy-efficient overall. Its disadvantage is that large detours may have to be accepted since the maximum acceleration has to fight against the prevailing mass inertia. These detours partially cancel out the efficiency advantage of a constant velocity.

Another path-based possibility to estimate flight times for UAV route planning problems is through the use of so-called Bézier curves \cite{Faigl.2018, Faigl.2019, Rodriguez.2017}. Here a start and end point are connected by a smooth curve, while the waypoints of a mission serve as control points to guide the curve. Since the resulting Bézier curve does not traverse all waypoints exactly, this method is suitable for solving so-called Close Enough Route Planning Problems (see \cite{Faigl.2019}), but it is less suitable when it is important to traverse all given waypoints exactly, as is the case for example in environments with obstacles.  

A third possibility for estimating the flight times, which at the same time guarantees a collision-free path with the environment, is described in \cite{Penicka.2019}. Here a collision-free path between the points to be visited is determined based on a probabilistic roadmap (PRM) algorithm. The path length is included as a cost in the overlaid route planning. With the objective of finding the fastest trajectory, using the path length as traversal costs might result in finding feasible trajectories, but these are most likely not the optimal ones.

All methods presented above, as well as most of the literature on UAV route planning, have in common that the traversal cost estimation is purely path-based (see \cite{Coutinho.2018}), which often entails the assumption of a constant flight velocity.

Investigating the field of trajectory optimization for UAVs, there is an abundance of approaches that enable UAVs to maneuver between two points in space by using a variable velocity. For example, minimum-snap trajectories are derived from higher-order polynomials in \cite{Mellinger.2011, Richter.2016, Liu.2017} and \cite{Burke.2020}. This usually results in a quadratic optimization problem whose solution reflects the coefficients of the polynomial trajectory.

Another possibility for trajectory optimization is the use of model-based predictive control (MPC). This is based on a discrete-time motion model of the UAV consisting of system parameters, system state, and control variables. To determine the flight trajectories, the deviation of the actual trajectory from the desired trajectory is penalized in a quadratic term. In addition, the use of control input such as acceleration or jerk is penalized quadratically as well. Again, this usually results in a quadratic optimization problem whose solution contains the system state as well as the control variable at each point in time and from which the flight motion of the UAV can be derived (see \cite{Mueller.2013, Konrad.2019}).

The last form of trajectory optimization presented here deals with the determination of time-optimal trajectories \cite{Beul.2016, Beul.2017} and is also based on polynomials. According to Pontryagin's Maximums Principle, time optimality is achieved by having the system always operate at the limit of the control range. For this purpose, the overall motion of the UAV is divided into several subsegments, all of which are described by a low-order polynomial. The control variable is constant in each of these subsegments and is either the maximum or minimum control value or is zero. By specifying start and end conditions, it is thus ultimately possible to calculate the time-optimal trajectory analytically.

Note that time optimality is also equal to energy optimality in terms of the total thrust integral. Faster trajectories might consume more energy for acceleration and deceleration, but since the overall flight time is shorter, their overall energy consumption is less compared to slower trajectories \cite{Beul.2016}.

\begin{figure}[h!]
	\centering
	\usetikzlibrary{shapes.misc}
\definecolor{b1}{RGB}{91,155,213}
\definecolor{b2}{RGB}{222,235,247}

\definecolor{r1}{RGB}{192,0,0}
\definecolor{r2}{RGB}{255,205,205}

\tikzstyle{circl2} = [circle, draw, fill = white,text width=0.4cm, text centered]
\tikzstyle{poly5} = [regular polygon, regular polygon sides = 5, draw, fill = white, text width=0.5cm, text centered, inner sep=0pt]
\tikzstyle{poly3} = [regular polygon, regular polygon sides = 3, draw, fill = white, text width=0.4cm, text centered, inner sep=0pt]
\tikzstyle{line1} = [draw = b1, -latex]
\tikzstyle{line2} = [draw = r1, -latex]

\begin{tikzpicture}[node distance = 1cm, auto, square/.style={regular polygon,regular polygon sides=4}]
	
%	\draw (-1.5,1.5) -- (1.5,1.5) -- (1.5, -1.5) -- (-1.5,-1.5) -- (-1.5,1.5);
	\node[circl2, line width = 1, minimum size=120] (one) {};
	\node at (0,0) [square,draw, inner sep= 30, fill=gray!30] (v100) {};
	\draw[line width = 1, ->] (-3,0) -- (3,0);
	\draw[line width = 1, ->] (0,-3) -- (0,3);
	\node[] at (3, -0.3) (x_axis) {$v_x, a_x$};
	\node[] at (-0.3, 3.2) (y_axis) {$v_y, a_y$};
	
	\draw[line width = 2, ->] (0,0) -- (0,1.5);
	\draw[line width = 2, ->] (0,0) -- (1.5,0);
	\draw[line width = 2, ->] (0,0) -- (1.5,1.5);
	\node[] at (-0.6, 0.75) (two) {$1/\sqrt{2}$};
	\node[] at (0.75, -0.4) (three) {$1/\sqrt{2}$};
	\node[] at (1.9, 2.2) (four) {$|v_{max}|, |a_{max}|$};
	\node[] at (1.2, 0.8) (four) {$1$};
\end{tikzpicture}
%}
	\caption{Division of the maximum velocity and acceleration in the plane to the individual coordinate axes by the factor $1/\sqrt{2}$. Although 
		all permissible velocity and acceleration vectors are inside the circle, only vectors inside the square can be realized by the decoupling process.}
	\label{fig:Kreis}
\end{figure}
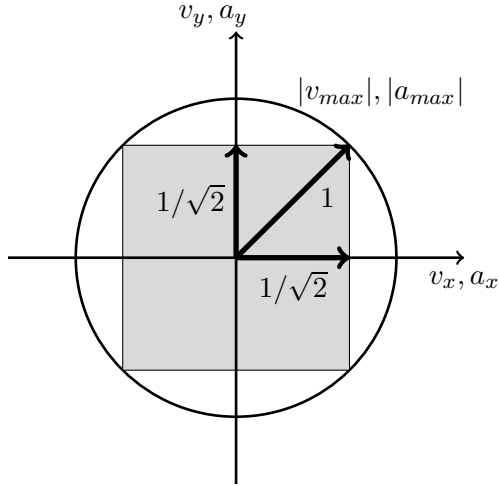

To simplify the trajectory planning problem, the $n$-dimensional motion is often decoupled into $n$ one-dimensional motions (see \cite{Faigl.2018, Hehn.2015} and \cite{Mueller2.2013}), but it is then important to note that any constraints on maximum velocity or acceleration are split among the $n$ partial trajectories by the appropriate factor. In a two-dimensional plane, for example, this factor is $1/\sqrt{2}$ which can be derived by the Pythagoras theorem  (see Figure \ref{fig:Kreis}).

Although an abundance of trajectory optimization methods seems to exist, approaches that combine trajectory optimization with UAV route planning are very sparse in literature \cite{Coutinho.2018}). To the best of our knowledge, there is no approach that solves the TBTSP as introduced in this paper.

\section{Routing}\label{Sec:Routing}
In this section, we present the optimization model for the Trajectory-based Traveling Salesman Problem. This model assumes that a UAV can move in a discretized direction through a waypoint and additionally at a discretized velocity. As a generalization of the TSP, the TBTSP is also NP-hard.
\subsection{Assumptions and Notations}
We solve a Traveling Salesman Problem for multirotor UAVs in an obstacle-free two-dimensional plane where each element in a set of waypoints $\mathcal{T} = \lbrace \mathcal{T}_i: i = 1,..., |\mathcal{T}|\rbrace$ must be visited with discretized heading angle and with a discretized velocity. Hence, $|\mathcal{T}|$ describes the number of tasks.

Following \cite{Cohen.2017}, we define the set of discretized heading angles as $\Theta = \left\{\theta_i: \theta_i = 2\pi i/|\Theta|,  i = 1,...,|\Theta|\right\}$, with $|\Theta|$ representing the number of discretization levels. The set of discretized velocities depends on the maximum allowed velocity and consists of elements $\mathcal{V} = \lbrace v_i: i = 1,..., |\mathcal{V}|\rbrace$ with $v_i \in \left[0, v_{max}\right]$. The number of discrete velocities is denoted by $|\mathcal{V}|$.  

Costs that describe the travel time to get from an initial waypoint $\mathcal{T}_i$ with heading angle $\theta_k$ and initial velocity $v_w$ to the final waypoint $\mathcal{T}_j$ with heading angle $\theta_m$ and final velocity $v_l$ are defined by $c_{ikw}^{jml}$.

\subsection{Mathematical Programming Formulation}
In this section, we present a mathematical programming formulation for our Trajectory-based Traveling Salesman Problem that provides the time-optimal trajectory to visit each waypoint with respect to heading angle and velocity. This formulation extends the work of \cite{Cohen.2017} by additionally considering discretized velocities for each waypoint.

The main decision variables for our formulation $x_{ikw}^{jml}$ are binary and interpreted as
\begin{equation*}
	x_{ikw}^{jml} = \begin{cases}
		1, & \text{if waypoint $i$ is left with heading angle $k$}\\
		& \text{and velocity $w$ towards waypoint $j$, which is }\\
		& \text{entered with heading angle $m$ and velocity $l$},\\
		0, & \text{otherwise}
	\end{cases}
\end{equation*}
Furthermore there are integer decision variables $u_i \in \left\{ 1, ..., |\mathcal{T}| \right\}, i=1,..., |\mathcal{T}|$
that define the sequence of waypoints $\mathcal{T}_i$ in the tour.

The resulting mathematical programming formulation for determining the optimal sequence of waypoints as well as the heading angle and velocity configurations with which to visit each waypoint is shown below.
\begin{align}
	\min & \sum_{i=1}^{|\mathcal{T}|}\sum_{j=1}^{|\mathcal{T}|} \sum_{k=1}^{|\Theta|}\sum_{m=1}^{|\Theta|}\sum_{w=1}^{|\mathcal{V}|} \sum_{l=1}^{|\mathcal{V}|} x_{ikw}^{jml}c_{ikw}^{jml} \label{TSP_objective}\\
	\text{s. t.} \nonumber\\
	& \sum_{j=1}^{|\mathcal{T}|} \sum_{k=1}^{|\Theta|}\sum_{m=1}^{|\Theta|}\sum_{w=1}^{|\mathcal{V}|} \sum_{l=1}^{|\mathcal{V}|} x_{ikw}^{jml} = 1, \quad i = 1,...,|\mathcal{T}| \label{each_task_entered}\\
	& \sum_{i=1}^{|\mathcal{T}|} \sum_{k=1}^{|\Theta|}\sum_{m=1}^{|\Theta|}\sum_{w=1}^{|\mathcal{V}|} \sum_{l=1}^{|\mathcal{V}|} x_{ikw}^{jml} = 1,\quad j = 1,...,|\mathcal{T}| \label{each_task_left}\\
	& \sum_{i=1}^{|\mathcal{T}|} \sum_{k=1}^{|\Theta|}\sum_{w=1}^{|\mathcal{V}|}x_{ikw}^{jml} - \sum_{o=1}^{|\mathcal{T}|} \sum_{p=1}^{|\Theta|}\sum_{q=1}^{|\mathcal{V}|}x_{jml}^{opq} = 0 \nonumber\\
	& \qquad\qquad\qquad\qquad\qquad\qquad  j=1,\cdots, |\mathcal{T}| \nonumber\\
	& \qquad\qquad\qquad\qquad\qquad\qquad  m=1,\cdots,|\Theta| \nonumber\\
	& \qquad\qquad\qquad\qquad\qquad\qquad  l=1,\cdots,|\mathcal{V}| \label{each_task_entered_is_left}
\end{align}
\begin{align}
	& u_i-u_j + (|\mathcal{T}|-1) \sum_{k=1}^{|\Theta|}\sum_{m=1}^{|\Theta|}\sum_{w=1}^{|\mathcal{V}|}\sum_{l=1}^{|\mathcal{V}|}x_{ikw}^{jml} \nonumber\\
	& + (|\mathcal{T}|-3) \sum_{k=1}^{|\Theta|}\sum_{m=1}^{|\Theta|}\sum_{w=1}^{|\mathcal{V}|}\sum_{l=1}^{|\mathcal{V}|}x_{ikw}^{jml} \leq |\mathcal{T}|-2 \nonumber\\
	& \qquad\qquad\qquad\qquad\qquad\qquad i=1,\cdots, |\mathcal{T}| \nonumber\\
	& \qquad\qquad\qquad\qquad\qquad\qquad j=1,\cdots, |\mathcal{T}| \label{subpath_prevention1}\\
	& u_1 = 1 \label{subpath_prevention2}\\
	%& 2 \leq u_i \leq |\mathcal{T}|, \,\qquad\qquad\qquad i = 2,...,|\mathcal{T}|			\label{subpath_prevention3}\\
	& u_i \in \lbrace 2, ..., |\mathcal{T}|\rbrace,\quad\qquad\qquad i = 2,...,|\mathcal{T}|  \label{subpath_prevention4}\\
	& x_{ikw}^{jml} \in \left\lbrace 0,1 \right\rbrace \qquad\qquad\qquad\,\,  i,j=1,\cdots, |\mathcal{T}| \nonumber\\
	& \qquad\qquad\qquad\qquad\qquad\qquad  k,m=1,\cdots,|\Theta| \nonumber\\
	& \qquad\qquad\qquad\qquad\qquad\qquad  w,l=1,\cdots,|\mathcal{V}| \label{binary}
\end{align}

The objective of the presented mathematical programming formulation \eqref{TSP_objective} is to minimize the total required travel time. Constraint set \eqref{each_task_entered} enforces that each waypoint is entered exactly once for all heading angles and velocities whereas constraint set \eqref{each_task_left} enforces that each waypoint is left exactly once. Flow conservation is fulfilled by constraint set \eqref{each_task_entered_is_left}. These constraints ensure that task $j$ is left in the same direction as it is entered as well as with the same velocity. To prevent subtours, we make use of the subtour elimination constraints \eqref{subpath_prevention1}, \eqref{subpath_prevention2}, and \eqref{subpath_prevention4}, which were developed in \cite{Desrochers.1991} and are an improved version of the Miller-Tucker-Zemlin Constraints \cite{Miller.1960}. Constraints \eqref{binary} enforce the decision variable $x_{ikw}^{jml}$ to be binary.

\section{Trajectory Generation} \label{Sec:TrajectoryGeneration}
To determine the cost $c_{ikw}^{jml}$ between two waypoints in the TBTSP model introduced in Section \ref{Sec:Routing}, we apply a simplified version of the trajectory generation approach presented in \cite{Beul.2016}. We assume the UAVs behaves like a point mass and can be accelerated in any direction with a maximum acceleration until a maximum velocity is reached. 

To use our approach, it is necessary to decouple the trajectories for each axis and optimize them separately. This procedure is well established in literature (see \cite{Faigl.2018, Hehn.2015, Mueller2.2013}) and is illustrated in Figure \ref{fig:Kreis}.

In the following, time-optimal trajectory optimization in one dimension is first described in more detail. This is followed by the description of the two-dimensional environment.

\subsection{One-dimensional Trajectory Generation}
Without loss of generality, we describe the trajectory generation for the $x$ axis in this section. Since our TBTSP formulation aims at time optimality, time optimality must also be aimed at in trajectory generation for the UAV. This is done by applying Pontryagin's Maximum Principle and operating the system at the limits of the control range. Furthermore, to guarantee feasibility in real world applications, the following constraints apply:
\begin{align}
	v_{x,min}\leq v_x \leq v_{x,max} \label{eq:constraint_v}\\
	a_{x,min}\leq a_x \leq a_{x,max} \label{eq:constraint_a}
\end{align}
Through these constraints, velocity $v_x$ and acceleration $a_x$ are restricted to their admissible range. In this work we define $v_{x,min} = -v_{x,max}$ and $a_{x,min} = - a_{x,max}$. The constraints \eqref{eq:constraint_v} and \eqref{eq:constraint_a} are further referred to as box constraints for the trajectory along the $x$ axis. 

According to Pontryagin's Maximum Principle and constraints \eqref{eq:constraint_v} and \eqref{eq:constraint_a}, we state that each time-optimal trajectory  consists of three phases of constant acceleration. Depending on the start and end state of the trajectory, we distinguish between four cases of different acceleration profiles (see Table \ref{tab:Cases}).  In cases 1 and 2 the first phase is associated with maximum and the last phase with minimum acceleration. In addition, we make a distinction between the direction in which the flight is to take place, which results in a change of sign for velocity and acceleration in cases 3 and 4. Further, we distinguish that the maximum velocity is reached once during the acceleration process (see cases 1 and 3) and once not (cases 2 and 4). 

\begin{table}[h]
	\centering\small
	\begin{tabular}{rrrr}
		\hline
		\addlinespace[1ex]
		\multicolumn{1}{c}{Case 1} & \multicolumn{1}{c}{Case 2} & \multicolumn{1}{c}{Case 3} & \multicolumn{1}{c}{Case 4} \\
		\addlinespace[1ex]
		\hline
		\addlinespace[1ex]
		$a_{x,0} = a_{x,max}$ & $a_{x,0} = a_{x,max}$ 	& $a_{x,0} = a_{x,min}$ & $a_{x,0} = a_{min}$ \\
		$a_{x,2} = a_{x,min}$ & $a_{x,2} = a_{x,min}$ 	& $a_{x,2} = a_{x,max}$ & $a_{x,2} = a_{max}$ \\
		$v_{x,1} = v_{x,max}$ & $t_{x,2} = 0$ 			& $v_{x,1} = v_{x,min}$ & $t_{x,2} = 0$\\
		\addlinespace[1ex]
		\hline		
	\end{tabular}
	\caption{Set of additional equations for each case.}
	\label{tab:Cases}
\end{table}

As an example, we explain case 1 from Table \ref{tab:Cases} in detail. During the first phase, there is a constant maximum acceleration until the maximum velocity is reached, followed by a second phase with constant maximum velocity and hence no acceleration. In the last phase, minimum acceleration, more specifically maximum acceleration in the inverse direction takes place.

In the following, $p_{x,i}$ describes the position and $v_{x,i}$ the velocity at the end of the $i$-th phase. Furthermore $t_{x,i}$ represents the duration of the $i$-th phase. The acceleration in the first, second and third phase is denoted by $a_{x,0}, a_{x,1}$ and $a_{x,2}$, respectively.

Moreover, we define starting and ending state for the $x$ axis by starting position $p_s$ and velocity $v_s$ and final position $p_e$ and velocity $v_e$. Note that $p_s, p_e, v_s, v_e$ result from the discretization in the TBTSP model and are given parameters. Since we set the start state to be active at the beginning of the first phase, the following equations hold.
\begin{align}
	p_{x,0} = p_s\label{eq:p0}\\
	v_{x,0} = v_s
\end{align}
The same holds for the end state, which is active at the end of the third phase.
\begin{align}
	p_{x,3} = p_e\\
	v_{x,3} = v_e \label{eq:v3}
\end{align}

Next, we define the end states of phases 1 - 3 by kinematic equations of motion which depend on the start state of the corresponding phase. 

\begin{align}
	p_{x,1} = p_{x,0} + v_{x,0} t_{x,1} + \frac{1}{2}a_{x,0}t_{x,1}^2 \label{eq:p1}\\
	p_{x,2} = p_{x,1} + v_{x,1} t_{x,2} + \frac{1}{2}a_{x,1}t_{x,2}^2 \\
	p_{x,3} = p_{x,2} + v_{x,2} t_{x,3} + \frac{1}{2}a_{x,2}t_{x,3}^2 \\
	v_{x,1} = v_{x,0} + a_{x,0}t_{x,1}\\
	v_{x,2} = v_{x,1} + a_{x,1}t_{x,2}\\
	v_{x,3} = v_{x,2} + a_{x,2}t_{x,3}
\end{align}

According to the maximum velocity constraint from equation \eqref{eq:constraint_v} the acceleration in the second phase is always equal to zero, since otherwise this constraint could be violated. Therefore, at the end of the first phase, the acceleration has to be shifted to zero and equation
\begin{align}
	a_{x,1} = 0 \label{eq:a2}
\end{align}
applies. 

When the equations \eqref{eq:p0} - \eqref{eq:a2} are combined with the equations of one specific case from Table \ref{tab:Cases}, a quadratic system of equations with 14 equations and 14 unknown variables results, which can be solved analytically. The duration $T_x$  of the trajectory along the $x$ axis containing all three phases is the sum of the duration of the individual phases. 
\begin{equation}
	T_{x} = t_{x,1} + t_{x,2} + t_{x,3}
\end{equation}

\subsection{Two-dimensional Trajectory Generation}
The duration of the entire trajectory in two dimensions is determined as the maximum of the durations over the trajectories for each axis.

\begin{equation}
	T_{traj} = \max_{i \in \left\lbrace x, y \right\rbrace} \left\lbrace T_i\right\rbrace  
\end{equation} 
The result enters the cost matrix for the TBTSP presented in Section \ref{Sec:Routing} as cost of connecting the corresponding waypoints. The procedure of using the maximum duration as cost for the TBTSP is based on the assumption the trajectory with the smaller duration always can be synchronized with $T_{traj}$.
\newline

\section{A computational study}\label{Sec:results}
To prove the concept of combining time-optimal trajectory generation with the TBTSP formulation proposed in Section \ref{Sec:Routing} we conduct a computational study. We benchmark our approach with the discretized Dubins Traveling Salesman Problem (DDTSP) presented in \cite{Cohen.2017}. Therefore, in the first step, we introduce the DDTSP and derive the parameters used. In the next step, we define the parameters used for our approach followed by presenting computational results.

For all calculations in this work, we used an Intel Core i7 - 8565U with a clock speed of 1.80GHz. The TBTSP model was implemented in Python 3.7 using Gurobi 9.1 as solver. The equations for our trajectory generation method were solved analytically using MAPLE 2021.   

\subsection{Benchmark Method}
As state-of-the-art approach for benchmarking we make use of a mathematical programming formulation of the discretized Dubins TSP (DDTSP) proposed in \cite{Cohen.2017}. To determine the edge costs for the DDTSP, we adapt the implementation from the python robotics toolbox \cite{Sakai.2018} to generate Dubins paths. We assume that the travel velocity is constantly equal to the maximum velocity allowed. The minimum turning radius, which serves as input for the implementation, is calculated by the maximum velocity $v_{max}$ and maximum acceleration $a_{max}$ according to
\begin{equation}
	R_{min} = \frac{v_{max}^2}{a_{max}}.
\end{equation}
The resulting Dubins path length $L$ is divided by the velocity $v_{max}$ to obtain the travel time $T_{Dubins} = L/v_{max}$. These travel times serve as costs for the DDTSP.

\subsection{Linking TBTSP and Trajectory Generation}
For our approach, we use the same parameters $v_{max}$ and $a_{max}$ as in the DDTSP for the box constraints
\begin{align}
	v_{min}\leq v_x \leq v_{max} \label{eq:constraint_v_tot}\\
	a_{min}\leq a_x \leq a_{max} \label{eq:constraint_a_tot}
\end{align}
of the entire two-dimensional trajectory. Note that we set $v_{min} = -v_{max}$ and $a_{min} = -a_{max}$. To ensure feasibility of the results, we also split $v_{max}$ and $a_{max}$ between the $x$ and $y$ axis (see Figure \ref{fig:Kreis}). Therefore, we set
\begin{align}
	v_{x,max} = v_{y,max} =  \frac{1}{\sqrt{2}} \cdot v_{max}, \\
	a_{x,max} = a_{y,max} =  \frac{1}{\sqrt{2}} \cdot a_{max}.
\end{align}
Furthermore, for each element $v_i \in \mathcal{V}$ it must hold that $v_i \in \left[0, \frac{1}{\sqrt{2}}\cdot v_{max}\right]$ to ensure feasibility, since otherwise the velocity to traverse a waypoint in two dimensions can be required to be higher than allowed by the box constraints from equations \eqref{eq:constraint_v} and \eqref{eq:constraint_a}. The initial and final velocities for each axis are then computed based on $v_i\in \mathcal{V}$ and $\theta_i \in \Theta$ as follows:
\begin{align}
	v_{x,i} = \sin(\theta_i)\cdot v_i \\
	v_{y,i} = \cos(\theta_i)\cdot v_i 
\end{align}

The initial and final acceleration in each waypoint is set to zero.
 
\subsection{Benchmark Instances}

To verify and demonstrate the potential of the method described in this paper, three problem instances are considered in more detail. The first instance contains nine waypoints which are regularly arranged in a 3x3 grid. The second instance contains 12 waypoints in a 3x4 grid and the third instance contains 16 waypoints in a 4x4 grid. The waypoints of each instance are located in one plane, which is why only movements in two spatial directions are considered. The shortest distance between two points is 9\,m. Thus, a multirotor UAV moving at a constant velocity of 1.5\,m/s and with a maximum lateral acceleration of 0.5\,m/s$^2$ as equivalent to a Dubins vehicle can just reach both waypoints by flying a semicircle.

For the computational study, the following cases are investigated. All instances are calculated with two different discretizations for the flight direction, once $|\Theta|=8$ and once $|\Theta|=16$. The discretization of $\Theta$ is always equidistant. Furthermore, for the DDTSP and our approach, the maximum velocities $v_{max}$ from the set $\lbrace 1.0, 1.5, 2.0, 2.5, 3.0 \rbrace$\,(m/s) are used. For the DDTSP the corresponding velocity is assumed as constant velocity. For the TBTSP, this is the maximum velocity, which finally still has to be divided among the $x$ and $y$ axis. Finally, we generate instances with two distinct velocity discretizations, which are $\mathcal{V} = \lbrace0.2, 0.6, 1.0\rbrace \cdot v_{max} \cdot 1/\sqrt{2}$ and $\mathcal{V}=\lbrace0.1, 0.2, ..., 1.0\rbrace \cdot v_{max} \cdot 1/\sqrt{2}$.

A maximum acceleration of $a_{max}=0.5$\,m/s$^2$ is assumed over all instances, which is composed of longitudinal and lateral component.

\subsection{Proof of concept}
The results of the study can be found in Tables \ref{tab:v3H8} to \ref{tab:v10H16}.  Table \ref{tab:v3H8} describes instances for which $|\mathcal{V}| = 3$ and $|\Theta| = 8$, Table \ref{tab:v10H8} instances with $|\mathcal{V}| = 10$ and $|\Theta| = 8$, Table \ref{tab:v3H16} instances with $|\mathcal{V}| = 3$ and $|\Theta| = 16$, and Table \ref{tab:v10H16} instances with $|\mathcal{V}| = 10$ and $|\Theta| = 16$. The table entries correspond to the obtained time of flight. The solution of the DDTSP is written in parentheses and is correspondingly the same for tables with an identical number of discretized flight directions. 

\begin{table}
	\centering
	\small
	\begin{tabular}{c|ccccc}
		\hline
		\addlinespace[1ex]
		Instance & 1.0 m/s & 1.5 m/s & 2.0 m/s & 2.5 m/s & 3.0 m/s \\
		\addlinespace[1ex]
		\hline
		\addlinespace[1ex]
		\multirow{2}{*}{3x3}&  (89.47) & (69.62) & (89.72) & (119.86) & (139.67)\\
							&   119.24 & 83.40   &  68.11  & 62.33    & 62.44 \\
		\addlinespace[1ex]
		
		\multirow{2}{*}{3x4}&  (110.91) & (79.21) & (100.53) & (122.16) & (151.81)\\
							&   154.28  & 104.14  &  84.77   & 76.04    & 76.07   \\
		\addlinespace[1ex]
	
		\multirow{2}{*}{4x4}&  (146.85) & (101.07) & (119.19) & (145.86) & (181.15)\\
							&   205.53  & 138.60   &  107.32  & 99.69    & 99.38   \\
		\addlinespace[1ex]
		\hline
	\end{tabular}
	\caption{Optimal objective function value for $|\mathcal{V} = 3|$ and $|\Theta| = 8$}
	\label{tab:v3H8}
\end{table}
\begin{table}
	\centering
	\small
	\begin{tabular}{c|ccccc}
		\hline
		\addlinespace[1ex]
		Instance & 1.0 m/s & 1.5 m/s & 2.0 m/s & 2.5 m/s & 3.0 m/s \\
		\addlinespace[1ex]
		\hline
		\addlinespace[1ex]
		\multirow{2}{*}{3x3}&  (89.47) & (69.62) & (89.72) & (119.86) & (139.67)\\
							&   119.24 & 83.40   & 68.11   & 62.33    & 62.44\\
		\addlinespace[1ex]
		
		\multirow{2}{*}{3x4}&  (110.91) & (79.21) & (100.53) & (122.16) & (151.81)\\
							&   154.28  & 104.14  &  83.54   & 75.49    & 71.86   \\
		\addlinespace[1ex]
		
		\multirow{2}{*}{4x4}&  (146.85) & (101.07) & (119.19) & (145.86) & (181.15)\\
							&   205.53  & 138.60   &  107.19  & 95.52    & 91.72   \\
		\addlinespace[1ex]
		\hline
	\end{tabular}
	\caption{Optimal objective function value for $|\mathcal{V}| = 10$ and $|\Theta| = 8$}
	\label{tab:v10H8}
\end{table}
In all setups for the Dubins paths, the shortest flight time occurs at a velocity of around 1.5\,m/s. This is due to the previously mentioned minimum distance between two points of 9\,m, at which a UAV can just connect both points by a semicircle. If the constant velocity of the UAV is higher, the flight time associated with the DDTSP solutions increases. In contrast, the flight time based on our method for visiting all waypoints of a problem instance decreases steadily with increasing maximum velocity. Furthermore, the tables show that increasingly finer discretizations of flight directions and velocity are associated with shorter flight times.
\begin{table}
	\centering
	\small
	\begin{tabular}{c|ccccc}
		\hline
		\addlinespace[1ex]
		Instance & 1.0 m/s & 1.5 m/s & 2.0 m/s & 2.5 m/s & 3.0 m/s \\
		\addlinespace[1ex]
		\hline
		\addlinespace[1ex]
		\multirow{2}{*}{3x3}&  (88.89) & (69.62) & (83.96) & (119.86) & (139.67)\\
							& 119.24   & 83.40   & 67.70   & 61.87    & 60.71 \\
		\addlinespace[1ex]
		
		\multirow{2}{*}{3x4}&  (110.91) & (78.25) & (91.01) & (122.16) & (151.81)\\
							&   154.12  & 103.90  &  81.79   & 74.91   & 70.99   \\
		\addlinespace[1ex]
		
		\multirow{2}{*}{4x4}&  (146.83) & (100.95) & (102.74) & (145.86) & (181.15)\\
							&   205.05  & 137.88   &  104.76  & 93.45   & 84.18   \\
		\addlinespace[1ex]
		\hline
	\end{tabular}
	\caption{Optimal objective function value for $|\mathcal{V}| = 3$ and $|\Theta| = 16$}
	\label{tab:v3H16}
\end{table}

\begin{table}
	\centering
	\small
	\begin{tabular}{c|ccccc}
		\hline
		\addlinespace[1ex]
		Instance & 1.0 m/s & 1.5 m/s & 2.0 m/s & 2.5 m/s & 3.0 m/s \\
		\addlinespace[1ex]
		\hline
		\addlinespace[1ex]
		\multirow{2}{*}{3x3}&  (88.89) & (69.62) & (83.96) & (119.86) & (139.67)\\
							&   119.24 &  83.39  & 67.21   &   61.34  & 58.73  \\
		\addlinespace[1ex]
		
		\multirow{2}{*}{3x4}&  (110.91) & (78.25) & (91.01) & (122.16) & (151.81)\\
							&   154.12  & 103.90  &  81.79   & 73.34    & 69.50   \\
		\addlinespace[1ex]
		
		\multirow{2}{*}{4x4}&  (146.83) & (100.95) & (102.74) & (145.86) & (181.15)\\
							&   205.05  & 137.87   &  104.76  & 90.62    & 83.21   \\
		\addlinespace[1ex]
		\hline
	\end{tabular}
	\caption{Optimal objective function value for $|\mathcal{V}| = 10$ and $|\Theta| = 16$}	
	\label{tab:v10H16}
\end{table}

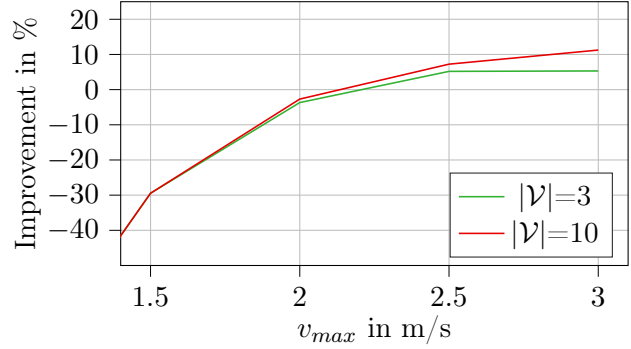
\begin{figure}[tb]
	\centering
	\begin{tikzpicture}
	\definecolor{mygreen}{rgb}{0.2,0.7,0.2}
	\definecolor{myred}{rgb}{0.9,0,0}
	
	\begin{axis}
		[
		width=3.25in,
		height=2in,
		tick align=outside,
		tick pos=left,
		xlabel={$v_{max}$ in m/s},
		xmin=1.4, xmax=3.1,
		xtick style={color=black},
		xtick={1.5,2, 2.5, 3},
		xmajorgrids,
		ymajorgrids,
		ylabel={Improvement in \%},
		ymin=-50, ymax=25,
		ytick style={color=black},
		ytick={-40, -30, -20, -10, 0, 10, 20},
		clip marker paths=true,
		legend style={at={(0.5, -0.0)}, anchor = north east, draw=white!15!black},
		legend pos = south east
		]		
		% |v| = 3
		\addplot [semithick, mygreen]
		table {%
			1.0 -89.80
			1.5 -29.46
			2.0 -3.68
			2.5 5.19
			3.0 5.31
		};
		\addlegendentry{|$\mathcal{V}$|=3}
		% |v| = 10
		\addplot [semithick, myred]
		table {%
			1.0 -89.80
			1.5 -29.46
			2.0 -2.69
			2.5 7.24
			3.0 11.25
		};
		\addlegendentry{|$\mathcal{V}$|=10}
		
%		\addplot [thick, black]
%		table {%
%			1.0 0
%			3.5 0
%		};
%		\addlegendentry{Dubins$^\ast$}
	\end{axis}
\end{tikzpicture}
	\vspace{-0.9cm}
	\caption{Travel time improvement of the TBTSP with $|\Theta| = 8$ compared with the optimal value achieved by the discretized Dubins TSP.}
	\label{fig:tikz_improvement_h8}
\end{figure}

\begin{figure}[tb!]
	\centering
	% This file was created by tikzplotlib v0.9.8.
\begin{tikzpicture}
	\definecolor{mygreen}{rgb}{0.2,0.7,0.2}
	\definecolor{myred}{rgb}{0.9,0,0}
	
	\begin{axis}
		[
		width=3.25in,
		height=2in,
		tick align=outside,
		tick pos=left,
		x grid style={white!69.0196078431373!black},
		xlabel={$v_{max}$ in m/s},
		xmajorgrids,
		xmin=1.4, xmax=3.1,
		xtick style={color=black},
		xtick={1.5,2, 2.5, 3},
		y grid style={white!69.0196078431373!black},
		ylabel={Improvement in \%},
		ymajorgrids,
		ymin=-50, ymax=25,
		ytick style={color=black},
		ytick={-40, -30, -20, -10, 0, 10, 20},
		clip marker paths=true,
		legend style={at={(0.5, -0.0)}, anchor = north east, draw=white!15!black},
		legend pos = south east
		]
		
		\addplot [semithick, mygreen]
		table {%
			1.0 -90.45
			1.5 -29.72
			2.0 -1.84
			2.5 7.61
			3.0 12.90
		};
		\addlegendentry{|$\mathcal{V}$|=3}
		
		\addplot [semithick, myred]
		table {%
			1.0 -90.45
			1.5 -29.72
			2.0 -1.60
			2.5 9.46
			3.0 14.8
		};
		\addlegendentry{|$\mathcal{V}$|=10}
		
%		\addplot [thick, black]
%		table {%
%			1.0 0
%			3.5 0
%		};
%		\addlegendentry{Dubins$^\ast$}
		
	\end{axis}
	
\end{tikzpicture}
	\vspace{-0.9cm}
	\caption{Travel time improvement of the TBTSP with $|\Theta| = 16$ compared with the optimal value achieved by the discretized Dubins TSP.}
	\label{fig:tikz_improvement_h16}
\end{figure}
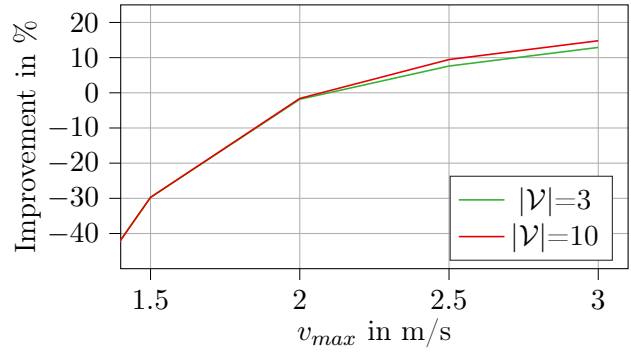

In Figures \ref{fig:tikz_improvement_h8} and \ref{fig:tikz_improvement_h16}, we give the average percentage deviation of the optimal solution of the TBTSP with different maximum allowed velocities relative to the best solution of the DDTSP at $v$\,=\,1.5\,m/s. It can be seen that  a maximum allowed velocity between 2\,m/s and 2.5\,m/s results in an improvement of up to 9\,\% relative to the best DDTSP solution. For a maximum allowed velocity of 3\,m/s, the average improvement compared to the best DDTSP solution even grows to just below 15\,\%. Note that the improvement for $|\mathcal{V}|=10$ is higher than the one achieved by $|\mathcal{V}|=3$. Nevertheless, both discretizations lead to a significant deterioration for low velocities, which is due to the fact that maximum allowed velocity and acceleration cannot be achieved except by moving diagonally between both axes (see Figure \ref{fig:Kreis}).

\begin{figure}[tb!]
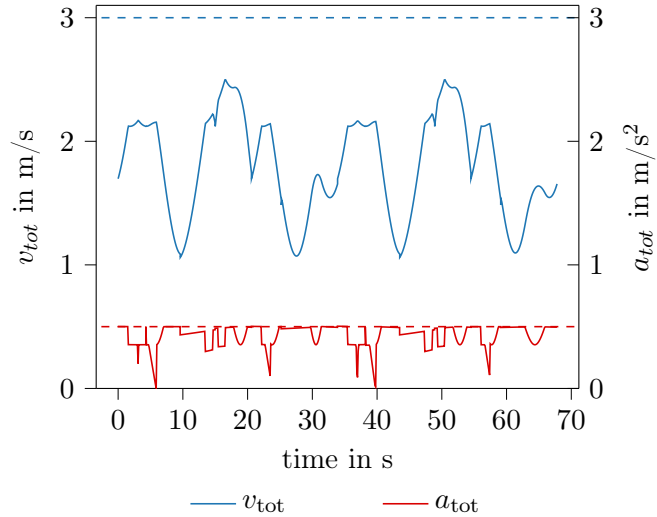

	\centering
	\include{tikz/velocity_acceleration}
	\vspace{-0.9cm}
	\caption{Velocity and acceleration aggregated over two dimensions for instance 3x4, |$\mathcal{V}$|=10, |$\Theta$|=16, $v_{\text{max}}$ = 3\,m/s and $a_{\text{max}}$ = 0.5\,m/s$^2$}
	\label{fig:velocity_accelecation}
\end{figure}

However, it should be noted that despite the allowed maximum velocity of 3\,m/s, the UAV does not reach this velocity (see Figure \ref{fig:velocity_accelecation}) and remains below about 2.5\,m/s. Moreover, it can be seen that not all acceleration is fully utilized. Both aspects are due to the fact that the physically allowed velocity and acceleration are divided equally between the $x$ and $y$ axis and thus only $1/\sqrt{2}$ of $v_{max}$ and $a_{max}$ respectively can be used for pure motion along one coordinate axis.

\begin{figure}[tb!]
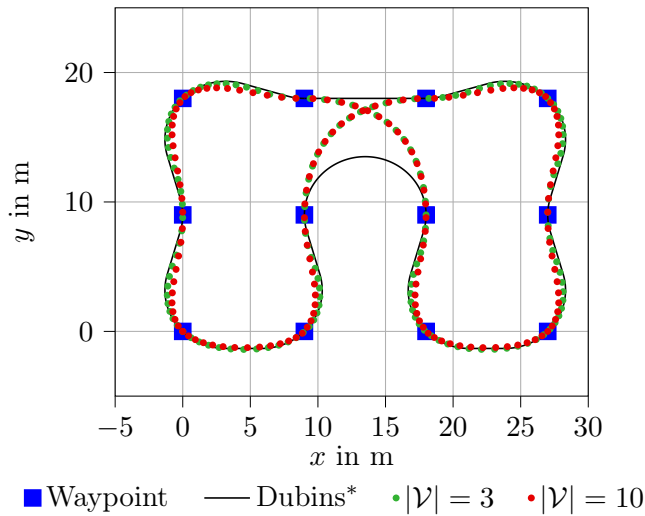

 	\centering
 	\include{tikz/Course_of_solutions_h8}
 	\vspace{-0.9cm}
 	\caption{Course of solutions for different configurations of $\mathcal{V}$, |$\Theta$|=8 fixed. It holds $v_{max}$=3\,m/s and $a_{max}$=0.5\,m/s$^2$ for TBTSP and $v_{max}$=1.5\,m/s for DDTSP.}
 	\label{fig:tikz_course_of_solutions_h8}
 \end{figure}

\begin{figure}[tb!]
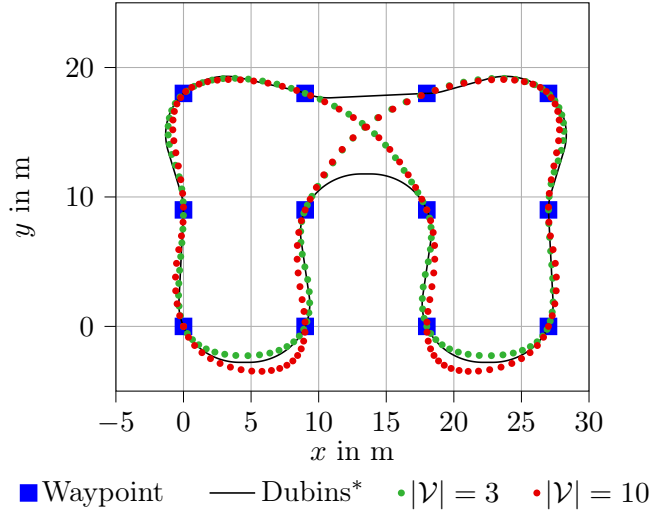

 	\centering
 	\include{tikz/Course_of_solutions_h16}
 	\caption{Course of solutions for different configurations of $\mathcal{V}$, |$\Theta$|=16 fixed. It holds $v_{max}$=3\,m/s and $a_{max}$=0.5\,m/s$^2$ for TBTSP and $v_{max}$=1.5\,m/s for DDTSP.}
 	\label{fig:tikz_course_of_solutions_h16}
 \end{figure}

The trajectory of the position with a maximum allowed velocity of 3\,m/s and different velocity discretizations can be seen in Figures \ref{fig:tikz_course_of_solutions_h8} and \ref{fig:tikz_course_of_solutions_h16}, where $|\Theta|= 8$ in the former and $|\Theta|=16$ in the latter. Also plotted in each figure is the best Dubins path, meaning at $v=1.5$\,m/s, for the corresponding directional discretization. 
 
 To synchronize the trajectories of each axis with the respective duration $T_{traj}$, we apply model predictive control (MPC) based trajectory generation, which is also subject to the box constraints from the equations \eqref{eq:constraint_v} and \eqref{eq:constraint_a} and uses acceleration as the control variable. 
 
 In the figures, the trajectories were sampled at a constant sampling rate, resulting in a dotted progression. Wide spacing between points therefore illustrates a high velocity whereas closely spaced points represent a low velocity. 

\subsection{Solution Times}
In this section, we first show the solution times to solve the TBTSP instances followed by a presentation of the solutions times of our trajectory generation method.

The average computation time to solve the TBTSP is given in Table \ref{tab:Solutiontimes}. The TBTSP optimization was stopped after an optimization time of 5000\,s. It can be seen that only the computation time of the 3x3 instances for |$\Theta$|=8 and |$\Theta$|=16 ran against this limit and were terminated prematurely. The optimality gap at termination for the 3x3 problem instance with $|\mathcal{V}=10|$ and $|\Theta|=16$ was highest and approximately 7.5\%. As the reason for instance 3x3 being the hardest to solve, we assume the point symmetry with respect to the middle waypoint.

\begin{table}\centering\small
	\begin{tabular}{crr|rr}
		\hline
		\addlinespace[1ex]
		&\multicolumn{2}{c|}{|$\mathcal{V}$|=3} & \multicolumn{2}{c}{|$\mathcal{V}$|=10}\\
		\addlinespace[1ex]
		\cline{2-3}  \cline{4-5}
		\addlinespace[1ex]
		Instance & |$\Theta$|=8 & |$\Theta$|=16 & |$\Theta$|=8 & |$\Theta$|=16  \\
		\addlinespace[1ex]
		\hline
		\addlinespace[1ex]
		3x3&   5.43  &  285.24 & 4434.40  & 5000.00  \\
		\addlinespace[2ex]
		3x4&   3.84  & 11.37   &  61.07   & 302.88   \\
		\addlinespace[2ex]
		4x4&   5.35  & 26.51   &  146.40  & 210.02   \\
		\addlinespace[1ex]
		\hline
	\end{tabular}
	\caption{Average solution times in seconds excluding trajectory generation}
	\label{tab:Solutiontimes}	
\end{table}

The computation of a single trajectory with the method presented in Section \ref{Sec:TrajectoryGeneration} takes about 22.8\,$\mu$s on average. The number of edges $\Omega$ to be calculated for solving the TBTSP is 
\begin{equation}
	\Omega= |\mathcal{T}|^2 \cdot |\Theta|^2 \cdot |\mathcal{V}|^2
\end{equation} 
For a problem instance with $|\mathcal{T}|=12$ waypoints, a flight direction discretization of $|\Theta|= 8$, and a velocity discretization of $|\mathcal{V}| = 10$, $\Omega= 921600$ trajectories result, all computed in a total of 21.00\,s.
 
\section{Conclusion and Outlook}\label{Sec:Conclusion}
This paper presents the TBTSP, a new extension to the TSP in which each waypoint can be traversed with discretized velocities and flight directions. In contrast to the state-of-the-art, which typically considers the UAV velocity to be constant, this model allows to explicitly maintain the degree of freedom of a variable velocity. 

To determine the edge cost, the trajectory generation of each axis is performed separately and the longest flight duration is ultimately used as the routing cost. The advantage of our trajectory planning approach is that it yields time-optimal trajectories, considers maximum velocity and acceleration constraints, computes feasible trajectories within microseconds and is significantly less complex than the approach presented in \cite{Beul.2016} while still being sufficiently precise in covering fundamental kinematic properties. 

The results show that the TBTSP finds better solutions than the DDTSP as state-of-the-art approach, as long as the maximum allowed velocity is high enough and waypoints are close to each other. However, a detailed evaluation of which problem instances benefit the most from our approach needs to be carried out in future research. 

Further performance potential arises from tuning the maximum acceleration and velocity of the individual axes, which has to be investigated in more detail in future research as well. 

Moreover, our exact solution approach shows the applicability of our model, but it is not practicable for larger problem instances with respect to computational effort. Therefore, a heuristic optimization method for solving the TBTSP will be developed with computational performance as the primary concern.
\section*{Acknowledgement}
Part of this research has been funded by the Federal Ministry of Education and Research of Germany.

\printbibliography

@article{Cohen.2017,
	author = {Cohen, I. and Epstein, C. and Shima, T.},
	year = {2017},
	title = {On the discretized {D}ubins {T}raveling {S}alesman {P}roblem},
	journal = {IISE Transactions},
	volume = {49},
	number = {2},
	pages = {238-254}
}

@article{Miller.1960,
	author = {Miller, C. E. and Tucker, A. W. and Zemlin, R. A.},
	year = {1960},
	title = {Integer programming formulation of {T}raveling {S}alesman {P}roblems},
	journal = {Journal of the ACM},
	volume = {7},
	pages = {326 - 329}
}

@article{Desrochers.1991,
	author = {Desrochers, M. and Laporte, G.}, 
	year = {1991},
	title = {Improvements and extensions to the {M}iller-{T}ucker-{Z}emlin subtour elimination constraints},
	journal = {Operations Research Letters},
	volume = {10}, 
	number = {1},
	pages = {27-36}
}

@article{Beul.2016, 
	author={Beul, M. and Behnke, S.},
	year={2016},
	title={Analytical time-optimal trajectory generation and control for multirotors},
	journal={International Conference on Unmanned Aircraft Systems (ICUAS)}
}

@article{Beul.2017, 
	author={Beul, M. and Behnke, S.},
	year={2017},
	title={Fast full state trajectory generation for multirotors},
	journal={International Conference on Unmanned Aircraft Systems (ICUAS)}
}

@article{Sakai.2018,
	author={Sakai, A. and Ingram, D. and Dinius, J. and Chawla, K. and Raffin A. and Paques, A.},
	year={2018},
	title={PythonRobotics: A {P}ython code collection of robotic algorithms},
	journal={Preprint arXiv:1808.10703}
}

@article{Mellinger.2011,
	author={Mellinger, D. and Kumar, V.}, 
	year={2011}, 
	title={Minimum snap trajectory generation and control for quadrotors},
	journal={IEEE International Conference on Robotics and Automation}
}

@article{Richter.2016,
	author={Charles, R. and Bry, A. and Roy, N.},
	year={2016},
	title={Polynomial trajectory planning for aggressive quadrotor flight in dense indoor environments},
	journal={Robotic Research},
	pages={649-666}
}

@article{Otto.2018,
	author={Otto, A. and Agatz, N. and Campbell, J. and Golden, B. and Pesch, E.},
	year={2018},
	title={Optimization approaches for civil applications of unmanned aerial vehicles ({UAV}s) or aerial drones: {A} survey},
	journal={Networks},
	volume={72},
	pages={411-458}
}

@article{Coutinho.2018,
	author={Coutinho, W. P. and Battarra, M. and Fliege, J.},
	year={2018},
	title={The unmanned aerial vehicle routing and trajectory optimization problem, a taxonomic review}, 
	journal={Computers and Industrial Engineering},
	volume={120},
	pages={116-128}	
}

@article{Mueller.2013,
	author={Mueller, M. W. and  D'Andrea, R.},
	year={2013},
	title={A model predictive controller for quadrotor state interception}, 
	journal={European Control Conference}	
}

@article{Konrad.2019,
	author={Konrad, T. and Salesch, T. and Abel, D.},
	year={2019},
	title={Flatness-based model predictive trajectory optimization for inspection tasks of multirotors},
	journal={American Control Conference}
}

@article{Hehn.2015,
	author={Hehn, M. and D'Andrea,R.},
	year={2015},
	title={Real-time trajectory generation for quadrocopters},
	journal={IEEE Transactions on Robotics},
	volume={31},
	number={4},
	pages={877-892}		
}

@article{Mueller2.2013,
	author={Mueller, M. W. and Hehn, M. and D'Andrea, R.},
	year={2013},
	title={A computationally efficient algorithm for state-to-state quadrocopter trajectory generation and feasibility verification},
	journal={International Conference on Intelligent Robots and Systems (IROS)}	
}

@article{Penicka.2017,
	author={Penicka, R. and Faigl, J. and Vana, P. and Saska. M.},
	year={2017},
	title={Dubins orienteering problem}, 
	journal={IEEE Robotics and Automation Letters},
	volume={2}, 
	number={2},
	pages={1210-1217}
}

@article{Penicka.2019,
	author={Penicka, R. and Faigl, J. and Saska. M.},
	year={2019},
	title={Physical orienteering problem for unmanned aerial vehicle data collection planning in environments with obstacles}, 
	journal={IEEE Robotics and Automation Letters},
	volume={4}, 
	number={3},
	pages={3005 - 3012}
}

@article{Geng.2013,
	author = {Geng, L. and Zhang, Y. F. and Wang, J. J. and Fuh, J. Y. H. and Teo, S. H.}, 
	year={2013}, 
	title={Cooperative task planning for multiple autonomous {UAV}s with graph representation and genetic algorithm}, 
	journal={IEEE Conference on Control and Automation}
}

@article{Faigl.2019,
	author = {Faigl, J. and Vana, P. and Penicka, R.},
	year = {2019},
	title={Multi-vehicle close enough orienteering problem with {B}ézier curves for multi-rotor aerial vehicles},
	journal={International Conference on Robotics and Automation}
}

@article{Faigl.2018,
	author = {Faigl, J. and Vana, P.},
	year = {2019},
	title={Surveillance planning with {B}ézier curves},
	journal={IEEE Robotics and Automation Letters},
	volume = {3},
	number={2},
	pages={750-757}
}

@article{Henchey.2020,
	author={Henchey, M. and Rosen, S.},
	year={2020},
	title={Emerging approaches to support dynamic mission planning: survey and recommendations for future research},
	journal={Journal of Defense Modelling and Simulation: Applications, Methodology, Technology}	
}

@article{Dubins.1957,
	author={Dubins, L. E.}, 
	year={1957},
	title={On curves of minimal length with a constraint on average curvature, and with prescribed initial and terminal positions and tangents},
	journal={American Journal of Mathematics}, 
	volume={79},
	number={3},
	pages={497-516}
}

@article{Burke.2020,
	author={Burke, D. and Chapman, A.  and Shames, I.},
	year={2020},
	title={Generating minimum snap quadrotor trajectories really Fast},
	journal={IEEE International Conference on Intelligent Robots and Systems (IROS)}
}

@article{Liu.2017,
	author={Liu, S. and Wattson, M. and Mohta, K. and Sun, K. and Bhattacharya, S. and C. T. Taylor, C. T. and Kumar, V.},
	year={2017},
	title={Planning dynamically feasible trajectories for quadrotors using safe flight corridors in 3-{D} complex environments},
	journal={IEEE Robotics and Automation Letters},
	volume={2},
	number={3},
	pages={1688-1695}
}

@article{Rodriguez.2017,
	author={Rodriguez, L. and Cobano, J. A. and Ollero, A.},
	year={2017},
	title={Smooth trajectory generation for wind field exploitation with a small {UAS}},
	journal={International Conference on Unmanned Aircraft Systems} 
}

@article{Beseda.2018,
	author={Beseda, J. A. and Bergesio, L. and Campana, I. and Vaquero-Melchor, D. and Araquistain, J. L. and Bernardos, A. M. and Casar, J. R.},
	year={2018},
	title={Drone mission definition and implementation for automated infrastructure inspection using airborne sensors},
	journal={Sensors},
	volume={18},
	number={4},
	pages={1170}
}

@article{DHL.2014,
	author={Heutger, M.  and Kueckelhaus, M.},
	year={2014},
	title={Unmanned aerial vehicle in logistics: A {DHL} perspective on implications and use cases for the logistics industry},
	journal={DHL Customer Solutions and Innovations}
}

@article{Shahmoradi.2020,
	author={Shahmoradi, J. and Talebi, E. and Roghanchi, P. and Hassanalian, M.},
	year={2020},
	title={A comprehensive review of applications of drone technology in the mining industry},
	journal={Drones},
	volume={4},
	pages={34}
}

@article{Zheng.2018,
	author={Zheng, X. and Wang, F. and Li, Z.},
	year={2018},
	title={A multi-{UAV} cooperative route planning methodology for 3{D} fine-resolution building model reconstruction},
	journal={ISPRS Journal of Photogrammetry and Remote Sensing},
	volume={146},
	pages={483-494}
}

@article{Mader.2016,
	author={Mader, D. and Blaskow, R. and Westfeld, P. and Weller, C.},
	year={2016},
	title={Potential of {UAV}-based laser scanner and multispectral camera data in building inspection},
	journal={The International Archives of Photogrammetry, Remote Sensing and Spatial Information Sciences},
	volume={41},
	pages={1135-1142}	
}

@article{Keller.2017,
	author={Keller, J. and Thakur, D. and Likhachev, M. and Gallier, J. and Kumar, V.},
	year={2017},
	title={Coordinated path planning for fixed-wing {UAS} conducting persistent surveillance missions},
	journal={Transactions on Automation Science and Engineering},
	volume={1},
	number={1},
	pages={17-24}
}

@article{Restas.2015,
	author={Restas, A.},
	year={2015},
	title={Drone applications for supporting disaster management},
	journal={World Journal of Engineering and Technology},
	volume={3},
	pages={316-321}
}

@article{Glock.2020,
	author={Glock, K. and Meyer, A.},
	year={2020},
	title={Mission planning for emergency rapid mapping with drones},
	journal={Transportation Science},
	volume={54},
	number={2},
	pages={534-560}
}

@article{Laporte.1992,
	author={Laporte, G.},
	year={1992},
	title={The {T}raveling {S}alesman {P}roblem: An overview of exact and approximate algorithms},
	journal={European Journal of Operational Research}, 
	volume={59},
	issue={2},
	pages={231-247}
}
\end{document}